\documentclass[sn-mathphys-num]{sn-jnl}

\usepackage{amsmath,amsfonts}
\usepackage{booktabs}
\usepackage{algorithmic}
\usepackage{graphicx}
\usepackage{textcomp}
\usepackage{algorithm}
\usepackage{xcolor}
\usepackage{hhline}
\usepackage{multirow}
\usepackage{tabularx}
\usepackage{subcaption}
\usepackage{threeparttable}

\begin{document}

\title{Deep Clustering via Distribution Learning}


\author[1]{\fnm{Guanfang} \sur{Dong}}\email{guanfang@ualberta.ca}
\equalcont{These authors contributed equally to this work.}

\author[1]{\fnm{Zijie} \sur{Tan}}\email{ztan4@ualberta.ca}
\equalcont{These authors contributed equally to this work.}

\author[1]{\fnm{Chenqiu} \sur{Zhao}}\email{chenqiu1@ualberta.ca}
\equalcont{These authors contributed equally to this work. \\This work is supported by NSERC Canada, and Alberta Innovates Discovery Supplement Grants}

\author[1]{\fnm{Anup} \sur{Basu}}\email{basu@ualberta.ca}

\affil[1]{\orgdiv{Department of Computing Science}, \orgname{University of Alberta}, \orgaddress{\street{116 St \& 85 Ave}, \city{Edmonton}, \postcode{T6G 2R3}, \state{Alberta}, \country{Canada}}}

%


\abstract{Distribution learning finds probability density functions from a set of data samples, 
	whereas clustering aims to group similar data points to form clusters. 
	Although there are deep clustering methods that employ distribution learning methods, past work still lacks theoretical analysis regarding the relationship between clustering and distribution learning.
	Thus, in this work, we provide a theoretical analysis to guide the optimization of clustering via distribution learning.
	To achieve better results, we embed deep clustering guided by a theoretical analysis.
	Furthermore, the distribution learning method cannot always be directly applied to data.
	To overcome this issue, we introduce a clustering-oriented distribution learning method called Monte-Carlo Marginalization for Clustering.
	We integrate Monte-Carlo Marginalization for Clustering into Deep Clustering, resulting in  Deep Clustering via Distribution Learning (DCDL).
	Eventually, the proposed DCDL achieves promising results compared to state-of-the-art methods on popular datasets.
	Considering a clustering task, the new distribution learning method outperforms previous methods as well.
	
	}




\maketitle

\section{Introduction}
Clustering is a fundamental task in the fields of data mining and computer vision \cite{zhou2022comprehensive}.
It involves grouping data points from a dataset into clusters, where data points within the same cluster exhibit high similarity.
While the optimization target seems straightforward, the design of an end-to-end clustering optimization method is not easy, especially considering high-dimensional data.
Thus, deep clustering is proposed by leveraging the fitting ability of deep neural networks to reduce the dimensionality of data, which achieves better results \cite{ren2022deep}.
With this motivation, we embed the concept of deep clustering in our proposed algorithm.

Given the dimension-reduced data, we still need a clustering algorithm to form clusters in an unsupervised manner.
In contrast to traditional clustering algorithms like k-means, distribution learning aims to learn the probability density functions from a set of data samples.
Although some existing methods embed distribution learning models such as Gaussian Mixture Model (GMM) in deep clustering, there is still a lack of theoretical analysis to support their relationship.
Besides, most distribution learning methods are not optimized for deep clustering.
This leads to a constrained search space for distribution learning algorithms, where the dimensionality of the data cannot be high.
Also, unoptimized algorithms may form imbalanced or meaningless clusters.

\begin{figure*}[tb]
	\includegraphics[width=\textwidth]{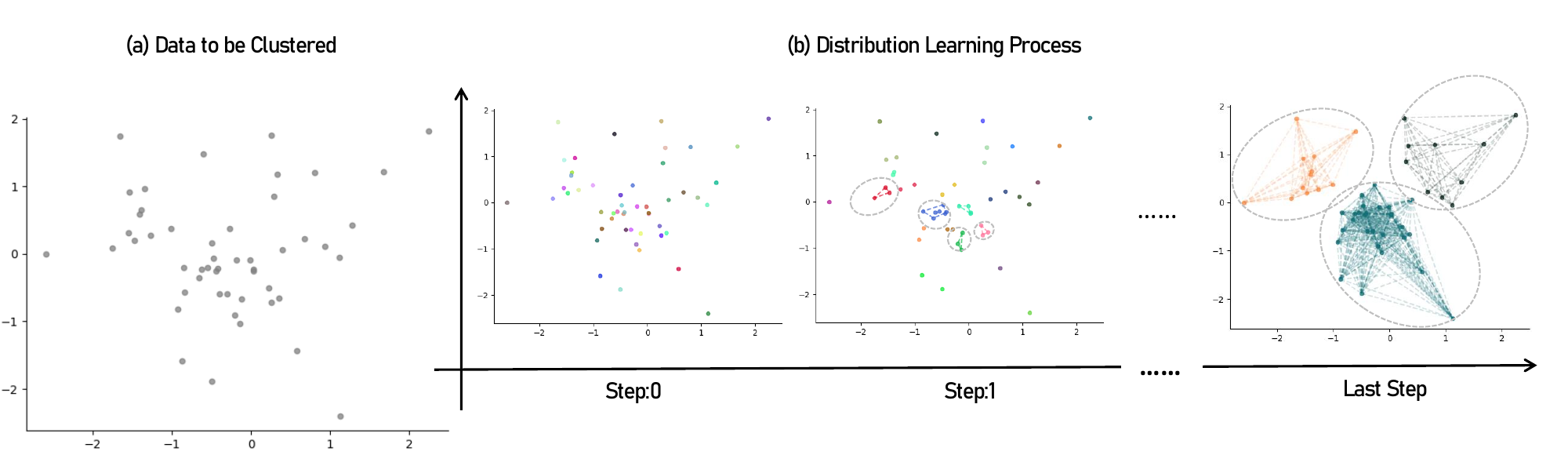}
	\caption{\label{fig1}
		\textbf{The Relationship between Clustering and Distribution Learning.} (a): Gray points represent the data to be clustered. (b): The process of distribution learning. We consider each data point is sampled from an underlying distribution, shown as Step 0 with each point possessing a distinct color. Then, to formulate an explicit expression of the distribution with cluster information, we redistribute the model components and align with the underlying prior distribution iteratively, as shown from Step 1 to the last step. This optimization objective aligns with clustering.}
\end{figure*}

The aforementioned limitations and the lack of theoretical foundations motivated us to propose Deep Clustering via Distribution Learning (DCDL). 
In DCDL, we first theoretically analyze the relationship between the clustering task and distribution learning.
As Figure \ref{fig1} shows, by treating each data point as a sample from an underlying distribution and considering the entire dataset as a mixture model, we can consider clustering as a process of simplifying a prior distribution.
Subfigure (a) demonstrates the data to be clustered.
Each point can represent a distribution component to form the initial mixture distribution, as shown in Subfigure (b) at Step 0.
However, the explicit expression from Step 0 is meaningless since it does not convey the clustering information. 
We need to redistribute to learn a meaningful distribution for the clustering task.
That is similar to compressing the prior distribution, as shown in the last step.

Following this concept, distribution learning can have the same optimization objective as clustering.
The clustering results achieved by distribution learning can now be supported by theory rather than relying solely on empirical observation.
Besides this, we propose a clustering-optimized distribution learning method called Monte-Carlo Marginalization for Clustering (MCMarg-C).
MCMarg-C is built on our previous work \cite{dong2023bridging}.
In MCMarg-C, we penalize excessively large or small clusters and initialize centers of clusters by prior guidance.
In addition, MCMarg-C can also directly learn distributions from very high dimensions (784 dimensions).
These features, along with the remarkable experimental results on popular datasets, suggest that our MCMarg-C may be one of the best distribution learning methods in clustering.

Compared to our previous work, ``\textbf{Bridging Distribution Learning and Image Clustering in High-dimensional Space} \cite{dong2023bridging}'', the contributions in this paper are:
\begin{enumerate}
	\item We conduct a theoretical analysis of the relationship between distribution learning and clustering.
	The analysis provides a novel perspective by viewing each data point as a distribution component.
	Thus, the distribution learning process can be seen as a redistribution of these Gaussian kernels.
	This aligns with the optimization objectives of clustering, providing theoretical support for using distribution learning in clustering problems.
	\item We introduce Deep Clustering via Distribution Learning (DCDL). 
	In DCDL, we integrate distribution learning into the deep clustering framework. 
	We employ an auto-encoder for dimensionality reduction and embed the latent vectors into a manifold space through manifold approximation. 
	Finally, we use the proposed Monte-Carlo Marginalization for Clustering (MCMarg-C) algorithm for distribution learning to obtain cluster labels. 
	Experimental results demonstrate that our DCDL outperforms state-of-the-art deep clustering methods.
	\item We propose a distribution learning method that is specifically designed for the clustering task, named Monte Carlo Marginalization for Clustering (MCMarg-C). 
	In MCMarg-C, we introduce prior guidance for means of cluster centers and penalize excessively large or small clusters.
	Considering that Monte Carlo Marginalization can also be used for extremely high dimensional data, MCMarg-C may be one of the best distribution learning methods in clustering. 
\end{enumerate}

\section{Related Work}
\label{related_work}

\subsection{Deep Clustering}
Due to the suboptimal performance of traditional clustering methods when applied directly to high-dimensional data, deep clustering methods have been proposed to map high-dimensional data into a feature space that is more suitable for clustering. 
In general, algorithms for deep clustering can be categorized into four main types: Deep Autoencoders (DAE) based algorithms \cite{song2013auto, chen2017unsupervised, huang2014deep, peng2016deep, xie2016unsupervised, guo2017improved, mcconville2021n2d, ghasedi2017deep, ren2020deep, li2018discriminatively, shah2017robust, yang2019deep2, affeldt2020spectral, yang2020adversarial}, Variational Autoencoders (VAE) based algorithms \cite{jiang2016variational, dilokthanakul2016deep, figueroa2017simple, li2018learning, willetts2019disentangling, prasad2020variational, cao2020simple}, Generative Adversarial Networks (GAN) based algorithms \cite{springenberg2015unsupervised, harchaoui2017deep, zhou2018deep, ghasedi2019balanced, mukherjee2019clustergan, mrabah2020adversarial, yang2022learning}, and Graph Neural Networks (GNN) based algorithms \cite{wang2019attributed, zhang2019attributed, tao2019adversarial, bo2020structural}. 

For DAE-based deep clustering methods, Huang et al. introduced Deep Embedding Network (DEN) \cite{huang2014deep} to enforce local constraints and group sparsity constraints on the learning objectives of DAE. 
Peng et al. \cite{peng2016deep} proposed deeP subspAce clusteRing with sparsiTY prior (PARTY) to enhance DAE with structural priors on the samples. 
Ren et al. introduced Deep Density-based Image Clustering (DDIC) \cite{ren2020deep} for density-based clustering on learned low-dimensional features. 
Xie et al. proposed Deep Embedding Clustering (DEC) \cite{xie2016unsupervised} inspired by t-Distributed Stochastic Neighbor Embedding (t-SNE) \cite{van2009learning}, which jointly optimizes the learning of features and clustering objectives. 
Guo et al. improved DEC and introduced Improved Deep Embedded Clustering (IDEC) \cite{guo2017improved}. 
Their improvements ensure local structure preservation during the fine-tuning phase of DEC. 
Subsequently, Guo et al. also introduced Deep Embedded Clustering with Data Augmentation (DEC-DA) \cite{guo2018deep} to enhance the performance of DEC using data augmentation. 

Recently, there has been a growing trend of DAE-based deep clustering methods with traditional machine learning algorithms. 
Affeldt et al. introduced Spectral Clustering via Ensemble Deep Autoencoder Learning (SC-EDAE) \cite{affeldt2020spectral} to integrate spectral clustering into DAE. 
Chen et al. introduced Deep Manifold Clustering (DMC) \cite{chen2017unsupervised}. 
They defined a locality-preserving objective to classify and parameterize unlabeled data points lying on multiple manifolds. 
Although Deep Clustering via Distribution Learning (DCDL) also embeds Uniform Manifold Approximation and Projection (UMAP), the only purpose is to map features to a suitable space for distribution learning.
In fact, manifold space transformation is a common practice in both deep and non-deep clustering methods \cite{souvenir2005manifold, elhamifar2011sparse, chen2017unsupervised, mcconville2021n2d}.

\subsection{Distribution Learning}
Distribution Learning focuses on finding the explicit expression for a given distribution. 
Although distribution learning seems straightforward, there are a limited number of methods for approximating distributions.
Kernel Density Estimation (KDE) \cite{rosenblatt1956remarks} estimates the Probability Density Function (PDF) by selecting an appropriate kernel function and computing the density.
Gaussian Mixture Model (GMM) \cite{bishop2006pattern} is a linear combination of multiple Gaussian distributions, and each component distribution has its mean, covariance, and mixture weights.
For GMM, the Expectation-Maximization (EM) \cite{dempster1977maximum} algorithm is frequently used to update the GMM parameters.
However, the EM algorithm suffers from challenges in handling high-dimensional data and non-differentiability, making it difficult to integrate with deep learning networks \cite{dong2023bridging, kingma2013auto}.
Thus, based on these needs, we proposed Monte-Carlo Marginalization (MCMarg) \cite{zhao2023learning} earlier, which is differentiable and can directly approximate high-dimensional data.
Also, some deep-learning distribution learning methods have been proposed.
Arithmetic Distribution Neural Network (ADNN) \cite{zhao2022universal} converts distributions to histograms and histogram distribution kernels are updated. 
Based on ADNN, some applications have also been proposed, including moving object segmentation \cite{dong2023learning}, vessel segmentation \cite{zhao2021pixel}, and affine-transformation-invariant image classification \cite{tan2023affine}.

\section{Methodology}
Our previous approach \cite{dong2023bridging} primarily validated the reliability of the connection between distribution learning and high-dimensional clustering problems.
The current work makes many adjustments to our existing distribution learning framework.
We provide a theoretical analysis that mathematically connects clustering and distribution learning.
We also introduce Deep Clustering via Distribution Learning (DCDL).
In this section, we first provide a theoretical analysis that explains the relationship between clustering and distribution learning. 
Subsequently, guided by the theoretical analysis, we demonstrate the pipeline and implementation details of DCDL.

\subsection{Problem Statement and Challenges}
The purpose of clustering is to divide data points into $k$ groups or clusters, such that points within the same cluster are similar to each other. 
However, due to the high dimensionality of images (a $28\times28$ image has 784 dimensions), the complexity of searching for and optimizing cluster regions may increase exponentially with an increase in the number of dimensions \cite{hammer1962adaptive}. 
Thus, as previously demonstrated in our earlier work, directly applying traditional clustering algorithms like K-means in a high-dimension yields suboptimal results. 
\par
Our previous work \cite{dong2023bridging} primarily explored the connection between the problems of clustering and distribution learning. 
Using Monte Carlo Marginalization (MCMarg) and direct observations in distribution learning, we significantly alleviated this issue, resulting in notable advantages over traditional clustering methods. 
Nonetheless, a formal analysis of the relationship between distribution learning and clustering is missing, which leads to inefficiencies when directly applying distribution learning methods to clustering problems. 
In light of this, we will next provide a theoretical review of the relationship between distribution learning and clustering and extend our previous work accordingly.

\subsection{Relationship between Distribution Learning and Clustering} \label{Theoretical}
\label{theoretical}
Given a clustering task in a multi-dimensional space, we want to give a theoretical analysis that bridges the problems of distribution learning and clustering. 
Let us start from the perspective of distribution learning and consider that all data points are sampled from an intractable distribution.
We first approximate the Probability Density Function (PDF) \(\mathcal{F}(\cdot)\) of this underlying target distribution\footnote{The target distribution is the distribution we want to learn. It means that this distribution has no explicit expression.} using Kernel Density Estimation (KDE) with Gaussian kernels.

Here, KDE places a kernel over each data point and computes the density estimate by adding these kernels.
This process can be seen as computing a mixture model of distributions whose probability density functions are kernel functions. 
Mathematically, given a multi-dimensional dataset \(\mathcal{X} = \{\mathbf{x}_i\} \subset \mathbb{R}^d\), the estimated probability density function \(\mathcal{F}(\cdot)\) is: 
\begin{equation}
	\begin{aligned}
		\mathcal{F}(x)= \frac{1}{N\cdot|H|^{\frac{1}{2}}} \sum_{i=1}^{N} \mathcal{K}(H^{-\frac{1}{2}}\cdot(\mathbf{x} - \mathbf{x}_i)),
	\end{aligned}
\end{equation} 
where \(N=|\mathcal{X}|\), \(\mathcal{K}(\cdot)\) is the multivariate kernel function, and \(H\) is a positive-definite \(d \times d\) bandwidth matrix. Since we use the Gaussian kernel:
\begin{equation}
	\mathcal{K}(\mathbf{x}) = \frac{1}{\sqrt{(2\pi)^d}} exp(-\frac{1}{2}\mathbf{x}^T\mathbf{x}), 
\end{equation}
we can rewrite \(\mathcal{F}(\cdot)\) as: 
\begin{equation}
	\label{target}
	\begin{aligned}
		\mathcal{F}(x)=&\frac{1}{N} \sum_{i=1}^{N} \frac{1}{\sqrt{(2\pi)^d|H|}} exp(-\frac{1}{2}(\mathbf{x} - \mathbf{x}_i)^TH^{-1}(\mathbf{x} - \mathbf{x}_i)) \\
		=&\sum_{i=1}^{N} \frac{1}{N} \mathcal{N}(\mathbf{x}; \mathbf{x}_i, H).
	\end{aligned}
\end{equation}
Compared to the PDF of a Gaussian Mixture Model (GMM):
\begin{equation} \label{GMM}
	p(\mathbf{x}) = \sum_{k=1}^{K} w_k \mathcal{N}(\mathbf{x}; \mathbf{\mu}_k, \mathbf{\Sigma}_k),
\end{equation}
where \(w_k\), $\mathbf{\mu}_k$ and $\mathbf{\Sigma}_k$ are weight vector, mean vector and covariance matrix. 
We can conclude that the estimated target distribution in Equation \ref{target} can be seen as another GMM that has \(N\) equal-weighted Gaussian components, each of which is centered at a unique data point \(x_i\), and uses the bandwidth \(H\) as its covariance matrix, as Figure \ref{fig1} (b), Step 0 shows. 
\par 
Thus, the optimization in a clustering problem can be reformulated from the viewpoint of a distribution learning problem.
That is, given the dataset \(\mathcal{X}\), we consider it as a mixture model of \(N\) underlying distributions, where each data point is sampled from the corresponding distribution component.
This underlying mixture model can be approximated by an equal-weighted GMM mirroring the entire dataset \(q(\mathbf{x})=\sum_{i=1}^{N} \frac{1}{N}\mathcal{N}(\mathbf{x}; \mathbf{x}_i, H)\).
We then need to optimize a set of parameters \(\Theta = \{(\mathbf{\mu}_k, \mathbf{\Sigma}_k)|k=1,\ldots,K\}\) for our GMM
\(p(\mathbf{x}; \Theta)=\sum_{k=1}^{K} w_k \mathcal{N}(\mathbf{x}; \mathbf{\mu}_k, \mathbf{\Sigma}_k)\), where \(K\) corresponds to the number of clusters. 
The optimization is guided by minimizing its divergence \(\mathcal{D}(\cdot)\) from the underlying distribution \(q(\mathbf{x})\).
Mathematically,
\begin{equation} \label{optimization}
	\Theta^*=\underset{\Theta}{\arg\min} \ \mathcal{D}(p(\mathbf{x};\Theta)||q(\mathbf{x})).
\end{equation}
\par		
Now, each data point $\mathbf{x}$ must be assigned to exactly one cluster, which is represented by a Gaussian component in our context. 
Hence, we can define a latent variable \(z_k\) such that \(z_k=1\) if the \(k\)-th cluster is selected; and \(z_k=0\) otherwise. Thus, \(\sum_k P(z_k=1)=1\). 
Now, we can formulate the process of assigning data points to clusters by finding the \(k\) that maximizes the probability of \(z_k=1\). 
That is, given the GMM parameter \(\Theta\), the cluster of data point \(\mathbf{x}\) is:
\begin{equation}
	\label{cluster1}
	\mathcal{C}(\mathbf{x}) = \underset{k}{\arg\max} \ P(z_k=1|\mathbf{x}, \Theta).
\end{equation} 
From Bayes' theorem, we have \(P(z_k=1|\mathbf{x}, \Theta)=\frac{P(\mathbf{x}|z_k=1, \Theta)P(z_k=1)}{P(\mathbf{x}|\Theta)}\). 
Now, since we have \(P(\mathbf{x}|\Theta) = \sum_{k} P(z_k=1)P(\mathbf{x}|z_k=1,\Theta) = \sum_{k} w_k\mathcal{N}(\mathbf{x}; \Theta_k)\), it can be inferred that the prior probability of \(P(z_k=1)\) can be represented by \(w_k\) in this scenario. 
Thus, Equation \ref{cluster1} can be rewritten as: 
\begin{equation}
	\mathcal{C}(\mathbf{x}) =  \underset{k}{\arg\max} \ \frac{w_k\mathcal{N}(\mathbf{x}; \Theta_k)}{\sum_{i} w_i\mathcal{N}(\mathbf{x}; \Theta_i)}.
\end{equation}
\par 
Additionally, to prevent any Gaussian component from having a significantly larger weight, we introduce a loss penalty term to minimize the variance of the weights of each Gaussian component. 
Specifically, we introduce GMM Weight Standard Deviation Loss (\( L_{\text{GMM-WSD}} \)):

\begin{equation} \label{loss_gmmwsd}
	\mathcal{L}_{\text{GMM-WSD}} = \sqrt{\frac{1}{K} \sum_{k=1}^{K} (w_k - \bar{w})^2}
\end{equation}
Where \(\bar{w}\) is the mean of weights in the learned GMM.
Since the standard deviation of the weights is penalized, the weights are stabilized. We can use the following Equation to determine the clusters:
\begin{equation}
	\label{cluster2}
	\begin{aligned}
		\mathcal{C}(\mathbf{x}) = \ & \underset{k}{\arg\max} \ \frac{w_k\mathcal{N}(\mathbf{x}; \Theta_k)}{\sum_{i} w_i\mathcal{N}(\mathbf{x}; \Theta_i)}\\
		\simeq \ & \underset{k}{\arg\max} \ \mathcal{N}(\mathbf{x}; \Theta_k).
	\end{aligned}
\end{equation}
This way, the clustering problem is reformulated as a distribution learning problem. 
\par
In practice, we use the monte-carlo marginalization method to approximate the Kullback-Leibler (KL) divergence and guide the optimization process. 
It first selects random unit vectors \( \vec{u} \), and marginalizes both the estimated target distribution \( q(\mathbf{x}) \) and our GMM \( p(\mathbf{x}; \Theta) \) along \( \vec{u} \) to obtain their respective marginal distributions, which are then compared in a lower-dimensional space. 
The optimization of the GMM parameters \( \Theta \) is guided by minimizing the KL divergence between these marginal distributions.
Mathematically, this is formulated as:
\begin{equation}
	\label{loss_kl}
	\begin{aligned}
		\mathcal{L}_{\text{KL}}(q(\mathbf{x}),p(\mathbf{x};\Theta))& =\int_{\vec{u}\in\mathbb{U}}\mathcal{D}_{KL}(q_{\vec{u}}(\mathbf{x}\cdot\vec{u})||p_{\vec{u}}(\mathbf{x}\cdot\vec{u};\Theta))d\vec{u}  \\
		&\simeq\sum_{\vec{u}\in\mathbb{U}}\mathcal{D}_{KL}(q_{\vec{u}}(\mathbf{x}\cdot\vec{u})||p_{\vec{u}}(\mathbf{x}\cdot\vec{u};\Theta)), 
	\end{aligned}
\end{equation}
where \(\mathcal{D}_{KL}(\cdot)\) is the KL divergence, and \(\mathbb{U} = \left\{\vec{u} \in \mathcal{M} \mid \|\vec{u}\|=1 \right\}\).
\par
Combining Equations \ref{loss_gmmwsd} and \ref{loss_kl}, we can now derive a complete objective function \(\mathcal{L}(\cdot)\). 
In our implementation, we also vectorize the computation of this objective function for efficiency as follows: 
\begin{equation}
	\label{loss_final}
	\begin{aligned}
		\mathcal{L}(\mathbf{x}, \mathbf{w}, \Theta) =&\ \mathcal{L}_{\text{KL}}(q(\mathbf{x}), \mathbf{w}^T\mathbf{\Psi}(\mathbf{x}; \Theta)) + c \times \mathcal{L}_{\text{GMM-WSD}}, \\
	\end{aligned}
\end{equation}
where \(\mathbf{w} = \left[w_1 \ w_2 \ \cdots \ w_K \right]^T\) is a vector consisting of weights in the GMM, \(c\) is a parameter that controls the impact of \(\mathcal{L}_{\text{GMM-WSD}}\), and \(\mathbf{\Psi}(\mathbf{x}; \Theta) = \left[ \mathcal{N}(\mathbf{x};\Theta_1) \ \ \mathcal{N}(\mathbf{x};\Theta_2) \ \ \cdots \ \ \mathcal{N}(\mathbf{x};\Theta_K) \right]^T\) is a vector-valued function, with each entry representing a Gaussian component of the GMM.
\par
Therefore, the learning and clustering can be expressed as: 
\begin{equation}
	\label{cluster3}
	\mathcal{C}(\mathbf{x}) = \underset{(k, \mathbf{w}, \Theta)}{\arg\max} \ \mathcal{L}(\mathbf{x}, \mathbf{w}, \Theta)^{-1}.
\end{equation}

\subsection{Implementation Details of Deep Clustering via Distribution Learning (DCDL)}
\begin{figure*}[tb]
	\includegraphics[width=\textwidth]{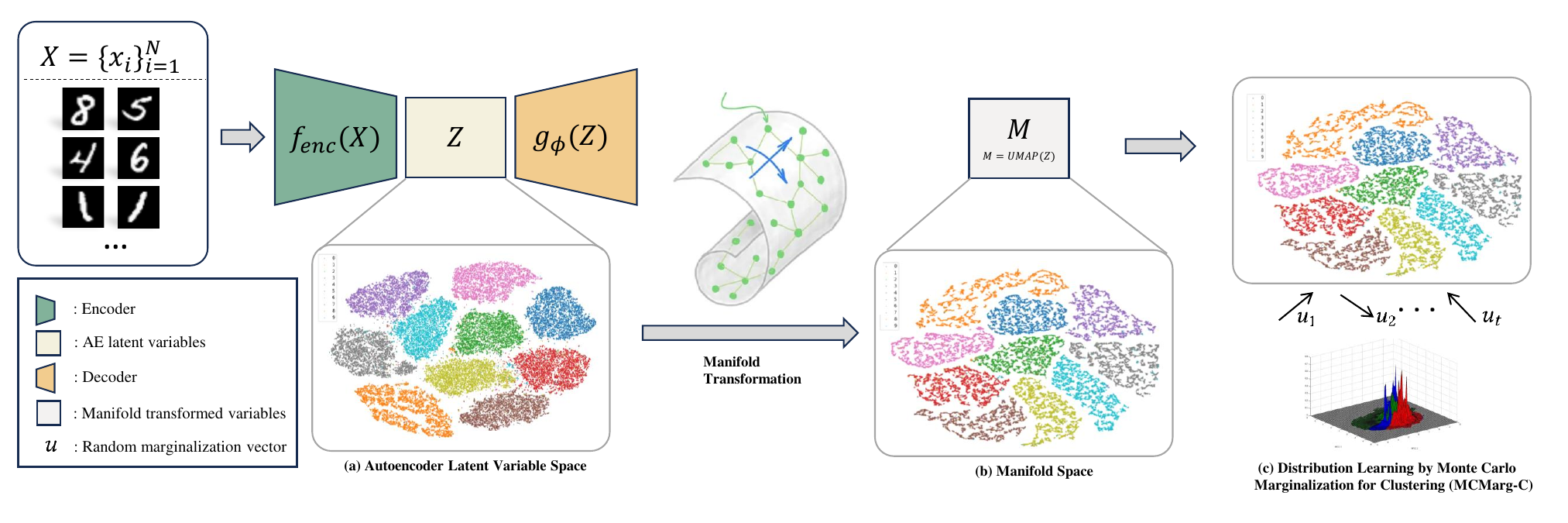}
	\caption{\label{deep_clustering}
		\textbf{Pipeline of Deep Clustering via Distribution Learning (DCDL).} The symbols depicted in the figure can be found with corresponding explanations in Algorithm \ref{algo:2}. Different colors in subfigures (a), (b), and (c) represent different labels in the MNIST dataset. The arrows in (c) represent the direction of marginalization in Monte Carlo Marginalization for Clustering (MCMarg-C).}
\end{figure*}
As depicted in Figure \ref{deep_clustering}, the proposed Deep Clustering via Distribution Learning (DCDL) demonstrates significant improvements and modifications compared to previous approaches. 
Consider a high-dimensional dataset \( \mathcal{X} = \{x_i\}_{i=1}^{N} \), where \( x_i \in \mathbb{R}^n \) represents a high-dimensional data point. 
Initially, dimensionality reduction is performed through a non-linear transformation by an Autoencoder, i.e., \( f_{\text{enc}}: \mathbb{R}^n \rightarrow \mathbb{R}^m \) where \( m \ll n \), to find a compact representation \( z_i = f_{\text{enc}}(x_i) \) for each high-dimensional data point \( x_i \). 
As this process involves deep neural networks, DCDL can also be categorized as a Deep Clustering algorithm.

Subsequently, all the low-dimensional encoded data \( \mathcal{Z} = \{z_i\}_{i=1}^{N} \) undergoes manifold approximation through Uniform Manifold Approximation and Projection (UMAP), denoted as \( f_{\text{umap}}: \mathbb{R}^{m} \rightarrow \mathbb{R}^{m^{\prime}} \). This maps the data to a manifold space \( \mathcal{M} = \{m^{\prime}_i\}_{i=1}^{N} \), where \( m^{\prime}_i = f_{\text{umap}}(z_i) \), to maintain the local and global structure of the data in the new space.

Following this theoretical analysis in Section \ref{theoretical}, we introduce an improved version of the MCMarg distribution learning algorithm, named MCMarg-C, specifically optimized for clustering tasks. 
The main enhancements in \textbf{MCMarg-C} include: (i) estimation of the Gaussian Mixture Model (GMM) means \( \mu_k \) through the k-means algorithm, i.e., \( \mu_k = \text{k-means}\ (\mathcal{M}) \), to enhance the independence between model components; 
(ii) the introduction of a Gaussian Mixture Model Weight Standard Deviation Loss (\( L_{\text{GMM-WSD}} \)), to achieve a balanced distribution of weights during the model learning process. This enhances the stability of the clustering results and avoids the bias of an overly dominant Gaussian component.

\subsection{Image Encoding}
The autoencoder (AE) is a commonly used method in deep clustering to reduce data dimension. 
AE consists of an encoder \( f_{\text{enc}}(x) \) and a decoder \( g_{\phi}(z) \); the encoder maps a high-dimensional data point into a latent vector like \( f_{\text{enc}}: \mathbb{R}^n \to \mathbb{R}^m \). The decoder performs the opposite operation, mapping from the latent space of dimension \( m \) back to the original high-dimensional space, as \( g_{\phi}(z) : \mathbb{R}^m \to \mathbb{R}^n \). Suppose we have \( N \) input data points in our high-dimensional dataset \( \mathcal{X} \); by applying the encoding function, we can obtain \( N \) points in the latent space, collected into a matrix \( \mathcal{Z} = \{z_i|i \in [1, N]\} \in \mathbb{R}^{N \times m} \). We approximate the distribution of \( \mathcal{Z} \) in the next step.

We use a simple autoencoder to perform feature embedding for high-dimensional data. 
The first layer reduces the dimension from $n$ to 500, followed by another layer maintaining the dimension at 500. 
Finally, we pass the data to a bottleneck layer that compresses the data into an $m$-dimensional latent space. 
The decoder part of the AE operates in the reverse direction. It takes the $m$-dimensional latent vector and gradually reconstructs the data back to its original dimensionality.

\subsection{Uniform Manifold Approximation and Projection (UMAP)}
In the previous section, our high-dimensional data \( \mathcal{X} \) was mapped to \( \mathcal{Z} = \{z_i|i \in [1, N]\} \in \mathbb{R}^{N \times m} \). 
A manifold is a hypothetical space where locally it resembles an Euclidean space, but globally it may have different shapes and structures. 
In our implementation, we chose to perform a Manifold Approximation of the embedded data obtained from the autoencoder using Uniform Manifold Approximation and Projection (UMAP) \cite{mcinnes2018umap}. 
Specifically, the UMAP algorithm consists of two steps. First, it reconstructs a neighborhood graph. 
In this step, for each point \( x_i \) in the dataset, UMAP determines its neighborhood size and calculates a distance metric, such as the Euclidean distance metric. 
Then, UMAP constructs a weighted graph $w$, where the weight of each point \( x_j \) in the local neighborhood of point \( x_i \) is given by:
\begin{equation}
	w_{ij} = \exp\left(-\frac{{d(x_i, x_j)^2}}{\sigma_i}\right) 
\end{equation}
Here, \( d(x_i, x_j) \) represents the distance between points \( x_i \) and \( x_j \), and \( \sigma_i \) is a parameter that adjusts the density of the local neighborhood.

Next, UMAP randomly selects some points in the manifold space for initialization. UMAP uses stochastic gradient descent to optimize the positions of points in the manifold space. 
The objective is to minimize the cross-entropy loss function, which quantifies the difference between the neighborhood graphs in the high-dimensional and manifold spaces. Specifically, the following loss function $C$ is applied:

\begin{equation}
	C = \sum_{i,j} w_{ij} \log\left(\frac{1}{1 + a||y_i - y_j||^{2b}}\right) + (1 - w_{ij}) \log\left(1 - \frac{1}{1 + a||y_i - y_j||^{2b}}\right) 
\end{equation}

Here, \( y_i \) and \( y_j \) represent points in the manifold space, and \( a \) and \( b \) are curve parameters learned from the data using a robust regression model.

\begin{figure*}[tb]
	\includegraphics[width=\textwidth]{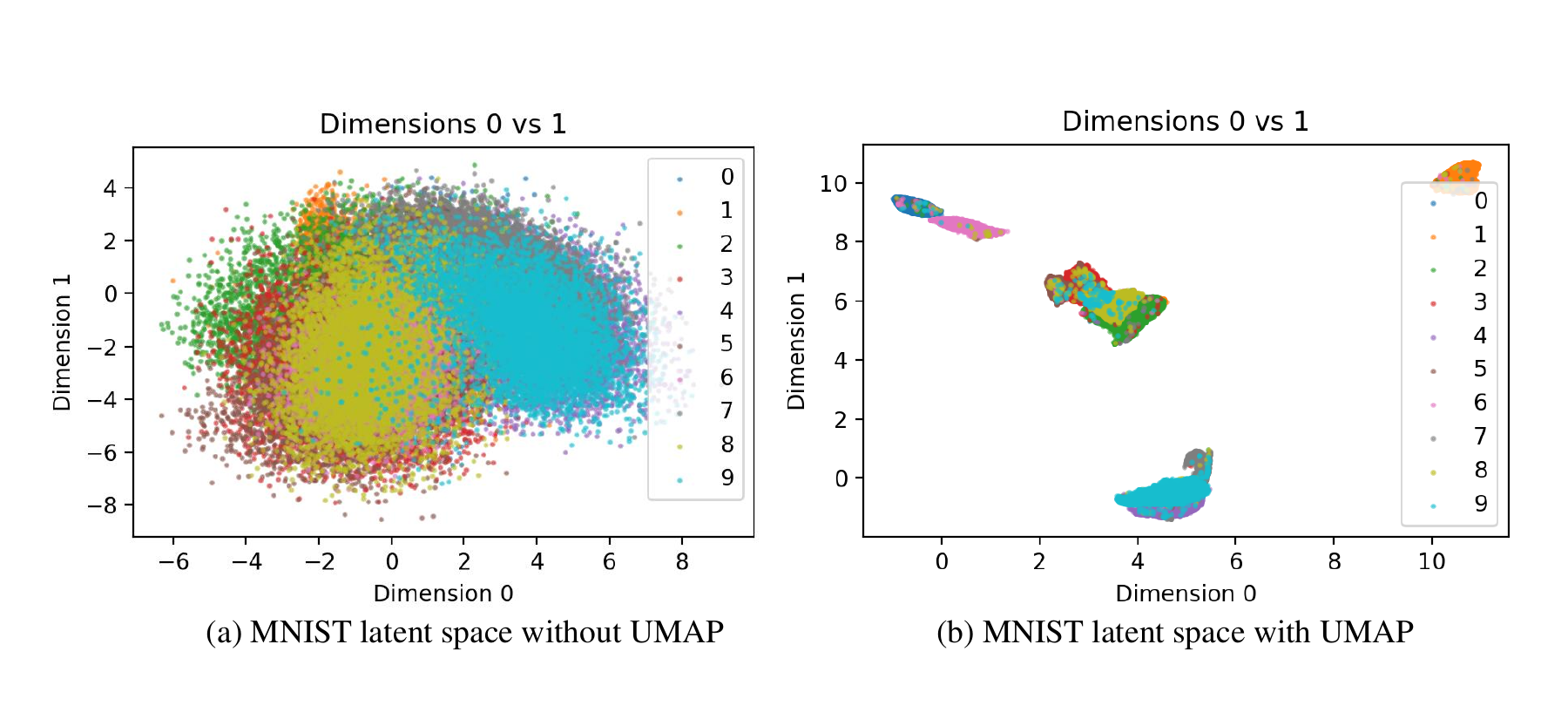}
	\caption{\label{mnist_no_umap}
		\textbf{Visualizing latent space of the MNIST dataset using autoencoder with and without using UMAP.} We visualize the plane projections of 0- and 1-dimensional spaces. We observe that the latent space transformed by UMAP exhibits sparser distributions between different labels and denser concentrations of points within each label.}
\end{figure*}

We visualize the embeddings of these two different spaces in Figure \ref{mnist_no_umap} for the MNIST dataset. 
The data distribution without transformation appears more disordered. 
However, the two-dimensional visualization results obtained with UMAP are more cohesive.

\subsection{Algorithm for Deep Clustering via Distribution Learning (DCDL)}
\begin{algorithm}[ht]
	\caption{Deep Clustering via Distribution Learning (DCDL) with \( n \) Clusters}
	\label{algo:2}
	\begin{algorithmic}[1]
		\REQUIRE High-dimensional data $\{x_i\}_{i=1}^N$, where $x_i \in \mathbb{R}^n$
		\ENSURE  Transformed representation $\{m^{\prime}_i\}_{i=1}^N$, GMM parameters $\{\theta_{\text{GMM}, k}\}_{k=1}^n$
		
		\STATE Initialize autoencoder parameters $\theta_{\text{enc}}$
		\FOR{each training iteration}
		\STATE Encode $x_i$ to get $z_i = f_{\text{enc}}(x_i)$
		\STATE Decode $z_i$ to reconstruct $\hat{x}_i = g_{\phi}(z_i)$
		\STATE Update $\theta_{\text{enc}}$ by minimizing $\sum_{i=1}^{N} \| x_i - \hat{x}_i \|^2$
		\ENDFOR
		
		\STATE Apply UMAP on encoded data $\{z_i\}_{i=1}^N$ to obtain $\{m^{\prime}_i\}_{i=1}^N$
		\STATE Initialize GMM parameters $\{\theta_{\text{GMM}, k}\}_{k=1}^n$ for \( n \) clusters
		\FOR{each transformed sample $m^{\prime}_i$}
		\FOR{each cluster $k = 1$ to $n$}
		\STATE Update parameters $\theta_{\text{GMM}, k}$ using MCMarg-C with $m^{\prime}_i$
		\STATE // The update can be represented as: $\theta_{\text{GMM}, k}^{(t+1)} = \text{MCMarg-C}\left( \theta_{\text{GMM}, k}^{(t)}, m^{\prime}_i \right)$
		\ENDFOR
		\ENDFOR
	\end{algorithmic}
\end{algorithm}
To better understand the proposed Deep Clustering via Distribution Learning (DCDL), we present the algorithm for DCDL in this section, namely Algorithm \ref{algo:2}.

\section{Experimental Results}
\subsection{Experimental Setting}

Our experiments are conducted on an NVIDIA RTX A4000 GPU. 
For the Auto-Encoder (AE), we utilize the Adam optimizer with a learning rate of 0.001 and apply batch normalization before generating the encoded vector.
For the Gaussian Mixture Model (GMM), we use the Adam optimizer with a learning rate of 0.0001, and the number of unit vectors sampled each time is 32.
For more experimental details, please refer to publicly available code\footnote{We will make public code available after the paper is published.}.

\subsection{Experimental Design}
We first conduct a qualitative and quantitative analysis of the clustering results for Deep Clustering by Distribution Learning (DCDL) on different datasets. 
Next, we compare DCDL with the traditional Gaussian Mixture Model updating algorithm, i.e., the Expectation-Maximization (EM) algorithm. 
Additionally, we present a qualitative and quantitative comparison between Monte Carlo Marginalization Clustering (MCMarg-C) and the previous MCMarg. 
Finally, we discuss the limitations of applying distribution learning in clustering problems and potential improvement strategies.\\
\textbf{Baseline Methods}:
Our initial motivation for designing DCDL is to achieve state-of-the-art results. 
With this motivation, we conduct comparisons with state-of-the-art deep clustering methods on popular datasets to demonstrate the effectiveness of DCDL. 
These methods include DeepCluster \cite{caron2018deep}, DCN \cite{yang2017towards}, IDEC \cite{guo2017improved}, SR-k-mean \cite{jabi2019deep}, VaDE \cite{jiang2016variational}, ClusterGAN \cite{mukherjee2019clustergan}, JULE \cite{yang2016joint}, DEPICT \cite{ghasedi2017deep}, and DBC \cite{li2018discriminatively}; they have top performance in the field of deep clustering.

Subsequently, a deep neural network encodes the high-dimensional image data into a low-dimensional space. 
This makes it possible to directly compare our distribution learning approach (MCMarg-C) with the Expectation-Maximization (EM) algorithm. 
Finally, we make a direct comparison with our previously proposed MCMarg method to demonstrate the superiority of MCMarg-C.\\
\textbf{Evaluation Metric}:
Our clustering performance is evaluated by three metrics: Adjusted Rand Index (ARI) \cite{warrens2022understanding}, Normalized Mutual Information (NMI) \cite{estevez2009normalized} and Top-1 Accuracy (ACC).

\textit{Adjusted Rand Index (ARI)} is a corrected version of the Rand Index (RI) and considers the effect of chance, making it suitable for evaluating the similarity between true and predicted cluster assignments.
The formula for calculating the Adjusted Rand Index takes into account the combinations of items within the clusters:
\begin{equation}
	ARI = \frac{{\sum\nolimits_{ij} \binom{n_{ij}}{2} - \left[\sum\nolimits_{i} \binom{a_i}{2} \sum\nolimits_{j} \binom{b_j}{2}\right] / \binom{n}{2}}}{{\frac{1}{2} \left[\sum\nolimits_{i} \binom{a_i}{2} + \sum\nolimits_{j} \binom{b_j}{2}\right] - \left[\sum\nolimits_{i} \binom{a_i}{2} \sum\nolimits_{j} \binom{b_j}{2}\right] / \binom{n}{2}}},
\end{equation}
where \( n_{ij} \) is the number of objects in both cluster \( i \) of the true clustering and cluster \( j \) of the predicted clustering. \( a_i \) is the sum of \( n_{ij} \) over all \( j \) for a fixed \( i \). \( b_j \) is the sum of \( n_{ij} \) over all \( i \) for a fixed \( j \).

\textit{Normalized Mutual Information (NMI)} is a statistical tool used to measure the similarity of clustering effects between two datasets. 
NMI is developed based on the concept of Mutual Information (MI) and involves normalization to ensure the evaluation is not affected by the size of clusters.
Here, Mutual Information is a measure of the mutual dependence between two random variables. 
The definition of Mutual Information is given by:
\begin{equation}
	MI(U, V) = \sum_{u \in U} \sum_{v \in V} P(u, v) \log\frac{P(u, v)}{P(u)P(v)},
\end{equation}
where \( U \) and \( V \) are two random variables, \( P(u, v) \) is their joint probability distribution, and \( P(u) \) and \( P(v) \) are their marginal probability distributions.
To overcome the issue of MI increasing with the number of clusters, normalization is introduced. Normalized Mutual Information (NMI) is achieved by dividing the MI value by a form of normalization, typically expressed as:
\begin{equation}
	NMI(U, V) = \frac{2 \times MI(U, V)}{H(U) + H(V)},
\end{equation}
where \( H(U) \) and \( H(V) \) are the entropies of the random variables \( U \) and \( V \) respectively. 
This normalization ensures that the NMI value lies between 0 and 1, where 0 indicates no correlation and 1 indicates perfect correlation.

\textit{Top-1 Accuracy (ACC)} is defined as the ratio of the number of times the clustering algorithm correctly predicts the most likely category to the total number of predictions made.
Mathematically, assume a dataset contains \( N \) samples. For each sample \( i \), we can generate a predicted category \( \hat{y}_i \) using the clustering algorithm and a true category \( y_i \). Top-1 Accuracy can be represented by the following formula:
\begin{equation}
	\text{Top-1 Accuracy} = \frac{1}{N} \sum_{i=1}^{N} 1(\hat{y}_i = y_i)
\end{equation}

Here, \(1(\hat{y}_i = y_i) \) is an indicator function that takes the value 1 when the clustered category \( \hat{y}_i \) equals the true category \( y_i \), and is 0 otherwise.

\subsection{Experimental Results}
\textbf{Comparison with State-of-the-Art Methods}

\begin{table}[h!]
	\centering
	\caption{\textbf{Comparison of different methods on MNIST, FashionMNIST, USPS, and Pendigits datasets.} Black bold represents the leading values, while red bold represents the second-ranked values.}
	\begin{tabular}{l|cc|cc|cc|cc}
		\toprule
		\textbf{Method} & \multicolumn{2}{c|}{\textbf{MNIST}} & \multicolumn{2}{c|}{\textbf{FashionMNIST}} & \multicolumn{2}{c|}{\textbf{USPS}} & \multicolumn{2}{c}{\textbf{Pendigits}} \\
		\hhline{~--------}
		& ACC         & NMI        & ACC          & NMI         & ACC        & NMI        & ACC         & NMI        \\ \midrule
		DeepGMM \cite{wang2021unsupervised}         & 0.7250      & 0.6400     & 0.4540       & 0.4100      & 0.6540     & 0.5100     & -           & -          \\
		DeepCluster \cite{caron2018deep}     & 0.7970      & 0.6610     & 0.5420       & 0.5100      & 0.5620     & 0.5400     & -           & -          \\
		DCN \cite{yang2017towards}             & 0.8300      & 0.8100     & 0.5010       & 0.5580      & 0.6880     & 0.6830     & 0.7200      & 0.6900     \\
		DEC \cite{xie2016unsupervised}            & 0.8630      & 0.8340     & 0.5180       & 0.5460      & 0.7620     & 0.7670     & 0.7010      & 0.6780     \\
		IDEC \cite{guo2017improved}           & 0.8810      & 0.8670     & 0.5290       & 0.5570      & 0.7610     & 0.7850     & 0.7840      & 0.7230     \\
		SC-EDAE \cite{affeldt2020spectral}        & 0.9323      & 0.8793     & -            & -           & 0.8178     & 0.8317     & \textcolor{red}{\textbf{0.8731}}      & \textcolor{red}{\textbf{0.8100}}     \\
		SR-k-means \cite{jabi2019deep}     & 0.9390      & 0.8660     & 0.5070       & 0.5480      & 0.9010     & 0.9120     & -           & -          \\
		VaDE \cite{jiang2016variational}           & 0.9450      & 0.8760     & 0.5780       & 0.6300      & 0.5660     & 0.5120     & -           & -          \\
		ClusterGAN \cite{mukherjee2019clustergan}     & 0.9640      & \textcolor{red}{\textbf{0.9210}}     & \textcolor{red}{\textbf{0.6300}}       & \textcolor{red}{\textbf{0.6400}}      & -          & -          & 0.7700      & 0.7300     \\
		JULE \cite{yang2016joint}           & 0.9640      & 0.9130     & 0.5630       & 0.6080      & \textcolor{red}{\textbf{0.9500}}     & \textcolor{red}{\textbf{0.9130}}     & -           & -          \\
		DBC \cite{li2018discriminatively}            & 0.9640      & 0.9170     & -            & -           & -          & -          & -           & -          \\
		DEPICT \cite{ghasedi2017deep}         & 0.9650      & 0.9170     & 0.5830       & 0.6200      & 0.8990     & 0.9060     & -           & -          \\
		DGG \cite{yang2019deep}            & \textbf{0.9760}      & 0.8800     & 0.6060       & 0.6100      & 0.9040     & 0.8200     & -           & -          \\ \midrule
		\textbf{DCDL (Our)}      & \textcolor{red}{\textbf{0.9722}}      & \textbf{0.9278}     & \textbf{0.6331}       & \textbf{0.6992}      & \textbf{0.9687}     & \textbf{0.9222}     & \textbf{0.8940}      & \textbf{0.8768}     \\ \bottomrule
	\end{tabular}
	\label{tab:method_comparison}
\end{table}

Table \ref{tab:method_comparison} compares the clustering results of DCDL with state-of-the-art methods on different datasets. 
Since these datasets come with labeled data, we can calculate accuracy (ACC) and Normalized Mutual Information (NMI). 
Quantitative comparisons based on ACC and NMI demonstrate the promising clustering performance of DCDL. 
As a deep clustering algorithm, DCDL achieves three first-place rankings in ACC and one second-place ranking. 
For NMI, DCDL secures first-place ranking across all datasets.

In particular, although both explicit and implicit distribution learning methods are indirectly used in deep clustering, they are both unable to deal with high-dimensional data and imbalanced clusters.
For example, DeepGMM \cite{wang2021unsupervised} learned the distribution explicitly via GMM.
The EM algorithm they used to update GMM is not specifically designed for the clustering task, which produces unsatisfactory results.
On the other hand, VaDE \cite{jiang2016variational} and ClusterGAN \cite{mukherjee2019clustergan}, which learn implicit distribution formation, also achieve suboptimal results.
VaDE utilizes Variational Autoencoder \cite{kingma2013auto} to map the data into the hypothetical distribution space.
However, the actual data distribution may not follow the distribution hypothesized by VAE, as shown by \cite{aneja2021contrastive, xiao2020vaebm, nalisnick2018deep, zhao2023learning}.
Thus, due to the differences between actual and presumed distributions, the clustering task does not perform well.
This drawback are also presented in GAN-based methods, where GAN \cite{goodfellow2020generative} adversarially updates generators to produce data that are close to the real data distribution.
For deep clustering, GAN may learn better data distributions than VAE, as ClusterGAN is better than VaDE in Table \ref{tab:method_comparison}.
However, MCMarg-C directly learns the explicit distribution without the concerns of high dimensionality, which helps MCMarg-C achieve better results than EM-based, VAE-based, and GAN-based deep clustering methods.

Moreover, for other deep clustering methods, we observe that DGG \cite{yang2019deep} achieves higher accuracy on MNIST (0.9760) compared to DCDL (0.9722). 
However, DGG's NMI score is significantly lower than ours (0.8800, compared to DCDL's 0.9278). 
This also validates that our algorithm can produce more balanced clustering results.
Compared with non-distribution learning deep clustering methods, these methods usually do not consider the formation of data distribution but form clusters based on other evidence, such as distance.
However, since we provide a theoretical analysis for distribution learning and design MCMarg-C that is more suitable for clustering, distribution learning may become a good choice for deep clustering.
The superior performance compared to non-distribution methods also verifies this.\\
\textbf{Comparison with Expectation-Maximization (EM) and Original Monte Carlo Marginalization (MCMarg)}

\begin{table}[t]
	\centering
	\caption{\textbf{Comparison of DCDL(EM), DCDL(MCMarg), and DCDL(MCMarg-C) on MNIST, FashionMNIST, USPS, and Pendigits datasets.} Bold values signify the top performance metrics across the datasets.}
\begin{tabular}{l|ccc|ccc}
	\toprule
	\textbf{Method} & \multicolumn{3}{c|}{\textbf{MNIST}} & \multicolumn{3}{c}{\textbf{FashionMNIST}} \\
	\hhline{~|---|---}
	& ACC & NMI & ARI & ACC & NMI & ARI \\
	\midrule
	DCDL(EM) & 0.9721 & 0.9276 & 0.9397 & 0.5899 & 0.6629 & 0.4668 \\
	DCDL(MCMarg) & 0.8331 & 0.8882 & 0.8188 & 0.5332 & 0.6521 & 0.4543 \\
	\textbf{DCDL(MCMarg-C)} & \textbf{0.9722} & \textbf{0.9278} & \textbf{0.9399} & \textbf{0.6331} & \textbf{0.6992} & \textbf{0.5207} \\
	\midrule
	\textbf{Method} & \multicolumn{3}{c|}{\textbf{USPS}} & \multicolumn{3}{c}{\textbf{Pendigits}} \\
	\hhline{~|---|---}
	& ACC & NMI & ARI & ACC & NMI & ARI \\
	\midrule
	DCDL(EM) & 0.9580 & 0.9012 & \textbf{0.9404} & 0.8928 & 0.8744 & 0.8070 \\
	DCDL(MCMarg) & 0.8834 & 0.8976 & 0.8654 & 0.7307 & 0.8024 & 0.6359 \\
	\textbf{DCDL(MCMarg-C)} & \textbf{0.9687} & \textbf{0.9222} & 0.9396 & \textbf{0.8940} & \textbf{0.8768} & \textbf{0.8090} \\
	\bottomrule
\end{tabular}
	\label{tab:integrated_comparison}
\end{table}

The Expectation-Maximization (EM) algorithm is an iterative optimization strategy used for estimating parameters in probabilistic models. 
The EM algorithm was introduced in the 1970s \cite{dempster1977maximum}. 
Over the following 50 years, although there have been numerous variations to the EM algorithm \cite{ng2012algorithm,liu1994ecme, fessler1994space,liu1998parameter,ng2003choice, gelfand1990sampling}, most of them have been based on theoretical innovations built upon the EM framework. 
No algorithm has managed to surpass the prominence of EM. 

In the E-step, based on the current parameter estimates, EM computes or estimates the expected values of hidden variables. Then, the M-step updates the parameter estimates to maximize the likelihood of the observed data. 
When applied to a high-dimensional space, the EM algorithm faces two primary challenges. 
First, the EM algorithm is very sensitive to initial values and may converge to a local optima in high-dimensional spaces. 
Second, the EM algorithm requires an update of the mean and covariance matrices for each Gaussian component during the M-step. 
Since this process has polynomial time complexity, the efficiency of the EM algorithm is significantly influenced by dimensionality, making its convergence difficult and time-consuming in high-dimensional spaces. 
Note that this is also one of the reasons why Variational Autoencoders (VAE) \cite{kingma2013auto} and Evidence Lower Bound (ELBO) were introduced. 
As mentioned in the original VAE paper (Section 2.1.1), \textit{``the EM algorithm cannot be used''} if the distribution is intractable.

Our previous work \cite{dong2023bridging} has demonstrated that for a high-dimensional distribution, MCMarg-C outperforms the EM algorithm.
Since a deep neural network is involved in reducing the data dimensionality, we create a condition favorable to the EM algorithm.
Thus, we also use the EM algorithm to update the GMM components and compare it with MCMarg-C. 
The comparison results are shown in Table \ref{tab:integrated_comparison}.

\begin{figure}[t]
	\centering
	\begin{subfigure}{.50\textwidth}
		\centering
		\includegraphics[width=\linewidth]{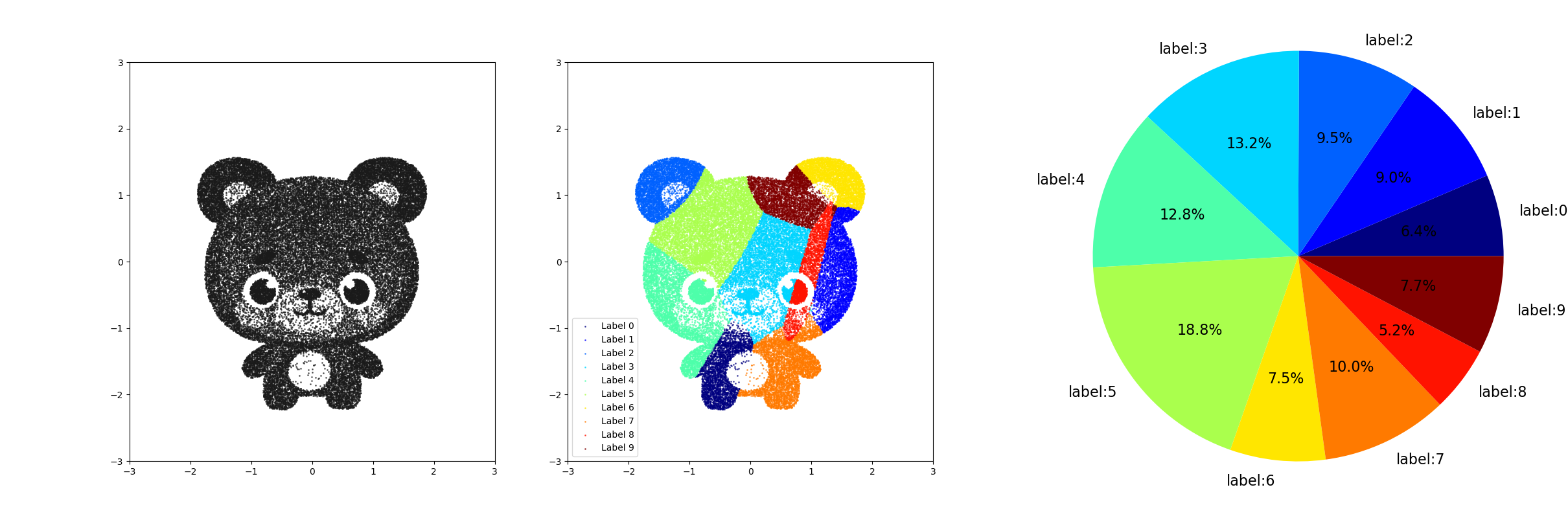}
		\caption{Bear Clustering Result by MCMarg}
		\label{fig:animal_0_o}
	\end{subfigure}%
	\hfill
	\begin{subfigure}{.50\textwidth}
		\centering
		\includegraphics[width=\linewidth]{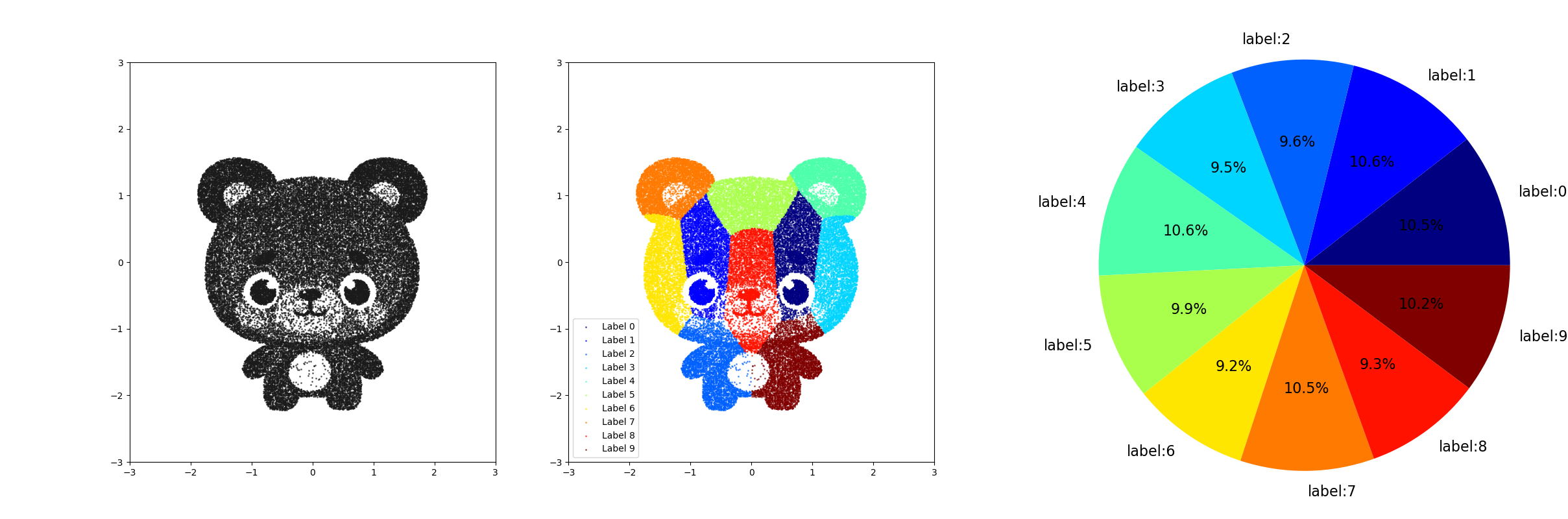}
		\caption{Bear Clustering Result by MCMarg-C}
		\label{fig:animal_0_c}
	\end{subfigure}
	
	\begin{subfigure}{.50\textwidth}
		\centering
		\includegraphics[width=\linewidth]{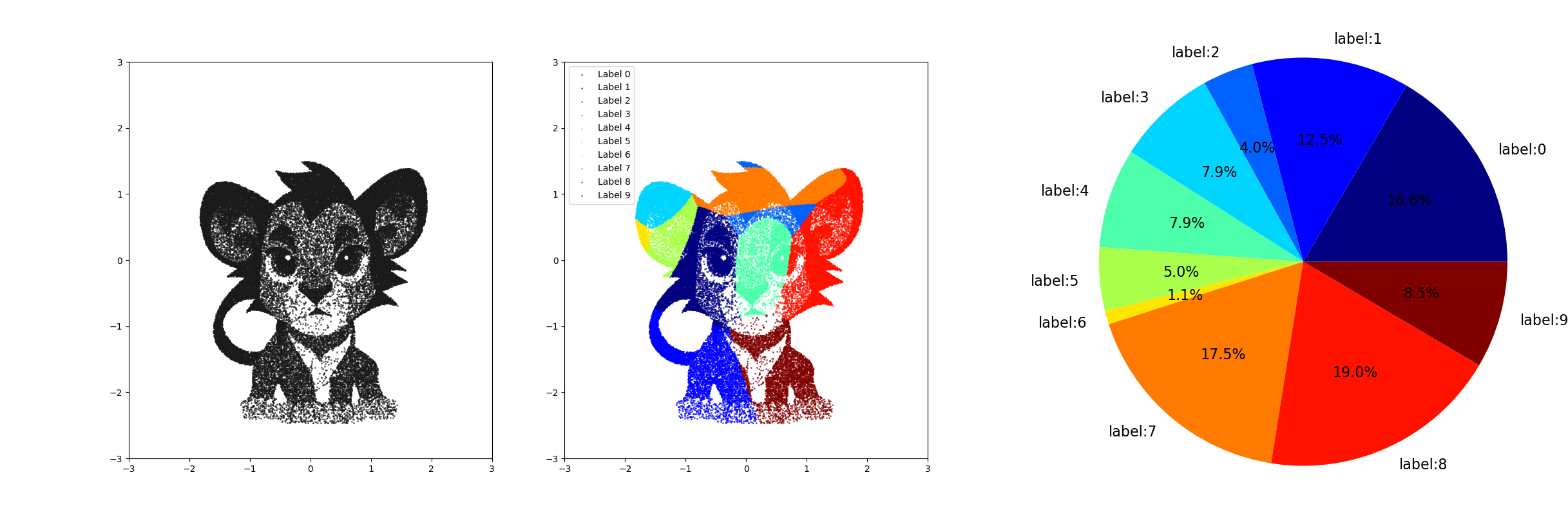}
		\caption{Lion Clustering Result by MCMarg}
		\label{fig:animal_6_o}
	\end{subfigure}%
	\hfill
	\begin{subfigure}{.50\textwidth}
		\centering
		\includegraphics[width=\linewidth]{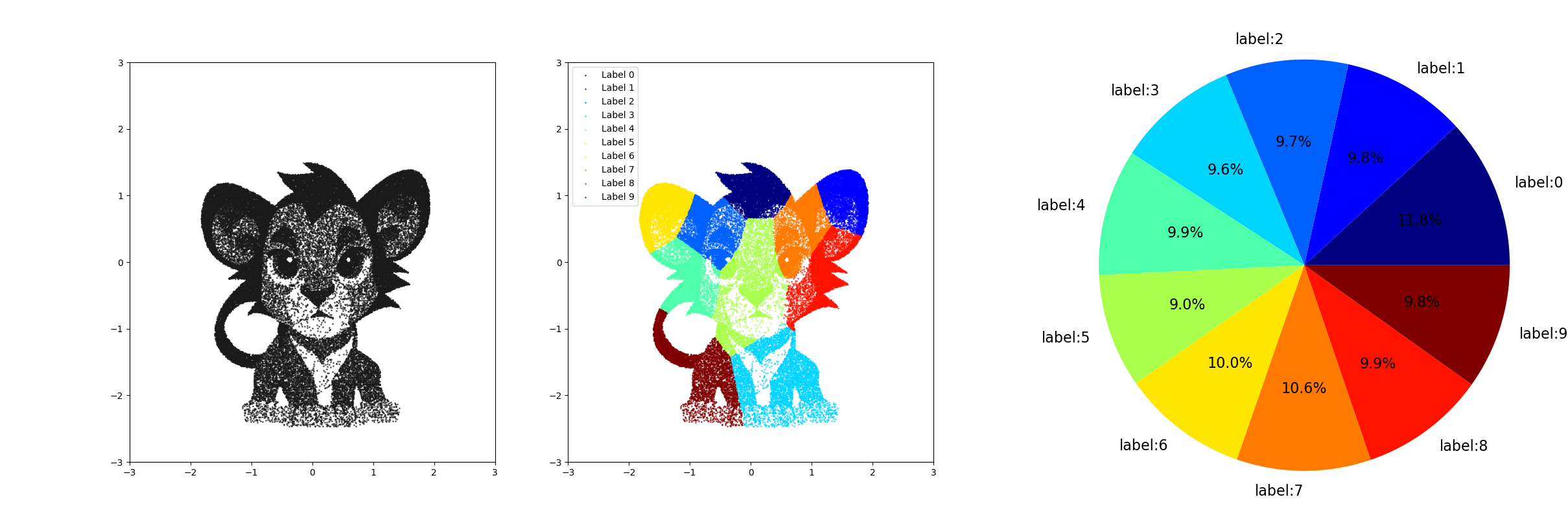}
		\caption{Lion Clustering Result by MCMarg-C}
		\label{fig:animal_6_c}
	\end{subfigure}
	
	\begin{subfigure}{.50\textwidth}
		\centering
		\includegraphics[width=\linewidth]{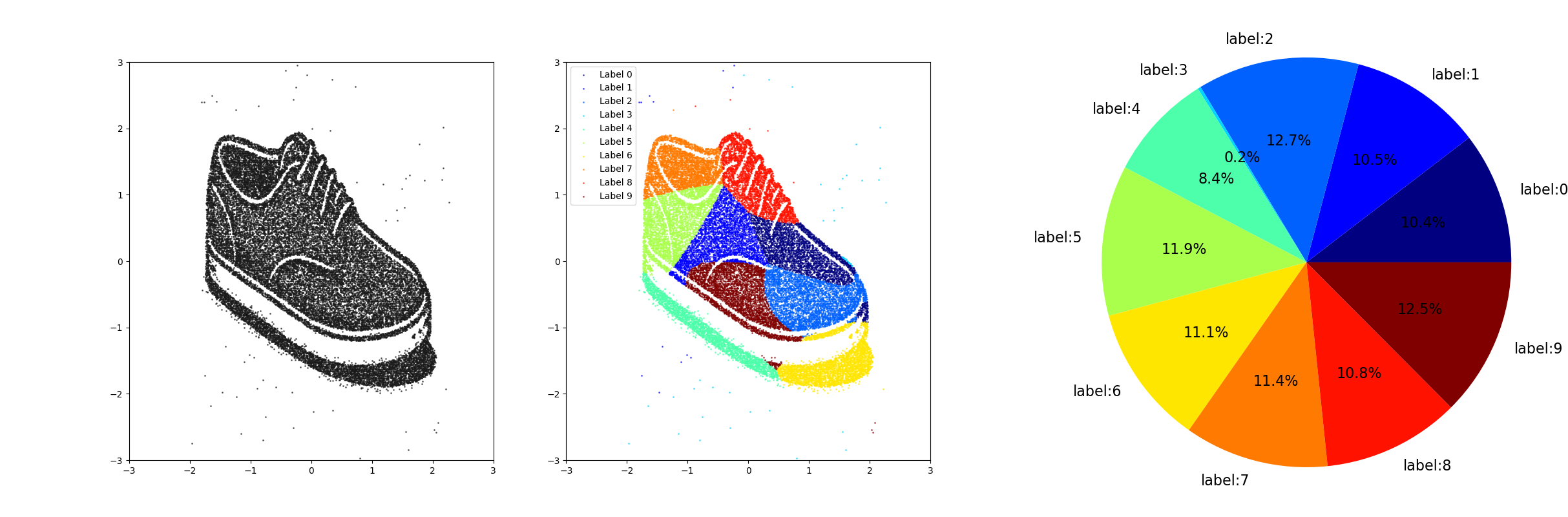}
		\caption{Shoe Clustering Result by MCMarg}
		\label{fig:shoe_5_o}
	\end{subfigure}%
	\hfill
	\begin{subfigure}{.50\textwidth}
		\centering
		\includegraphics[width=\linewidth]{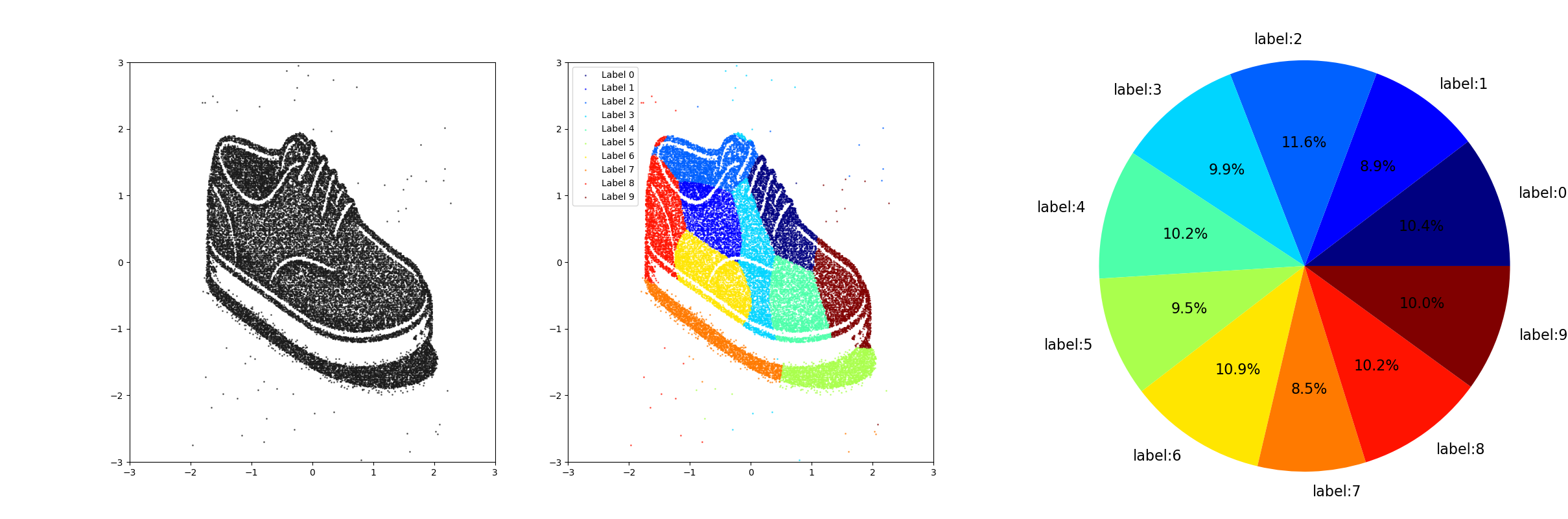}
		\caption{Shoe Clustering Result by MCMarg-C}
		\label{fig:shoe_5_c}
	\end{subfigure}
	
	\begin{subfigure}{.50\textwidth}
		\centering
		\includegraphics[width=\linewidth]{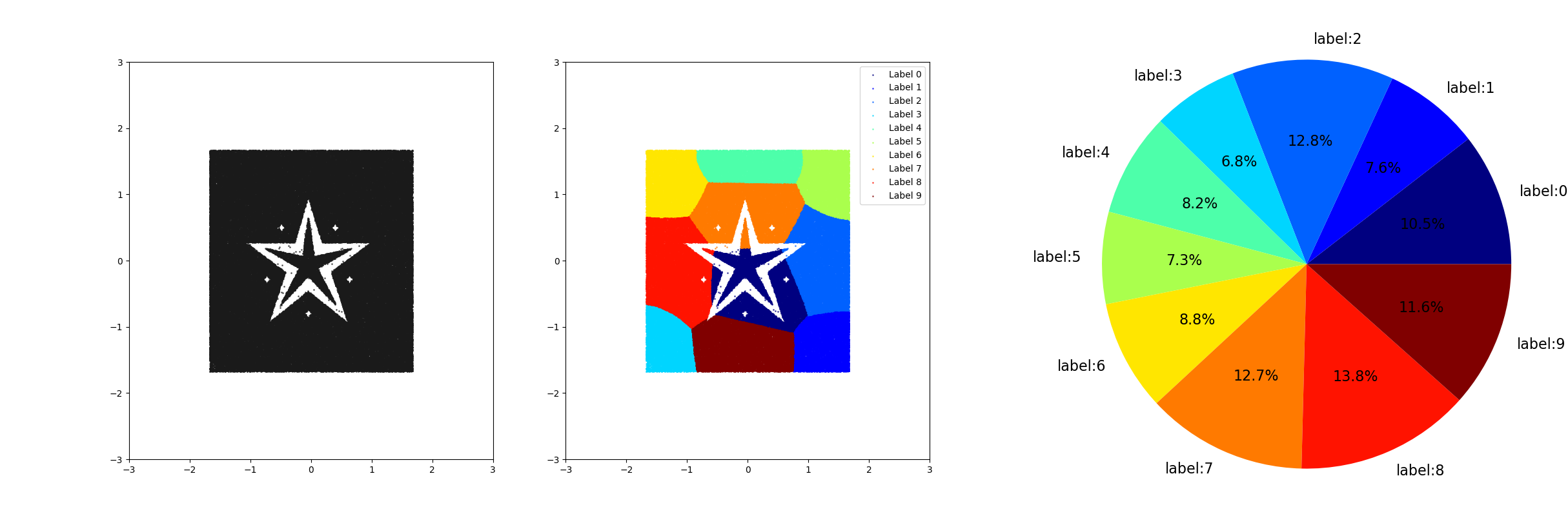}
		\caption{Star Clustering Result by MCMarg}
		\label{fig:star_5_o}
	\end{subfigure}%
	\hfill
	\begin{subfigure}{.50\textwidth}
		\centering
		\includegraphics[width=\linewidth]{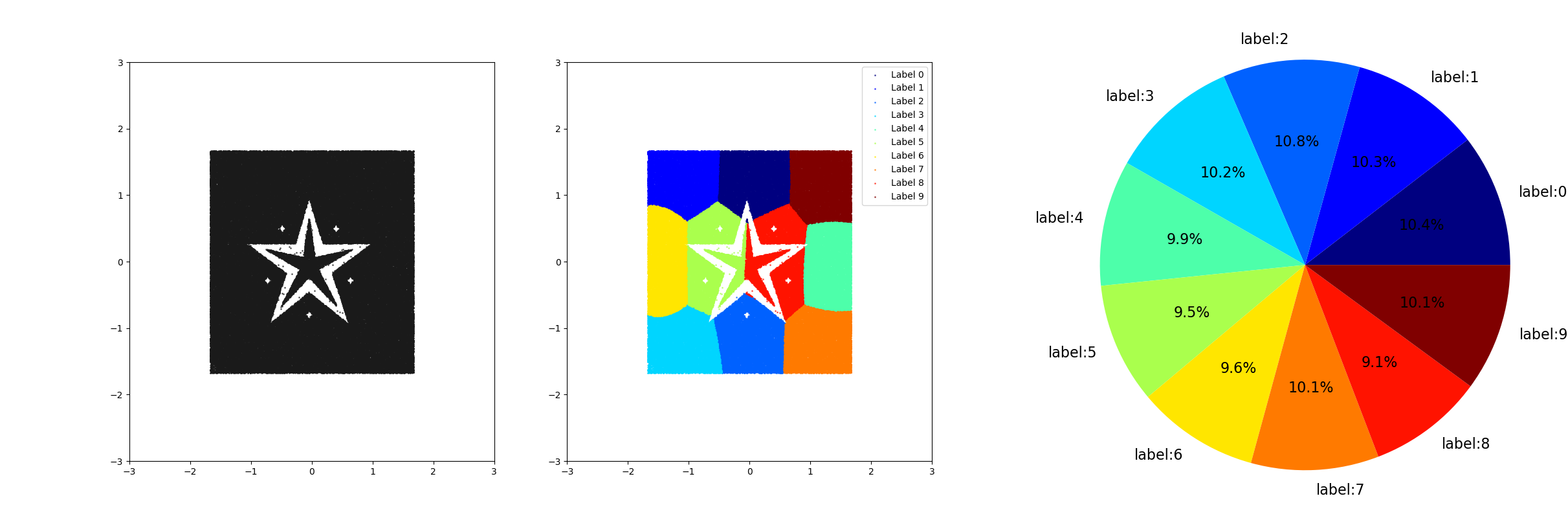}
		\caption{Star Clustering Result by MCMarg-C}
		\label{fig:star_5_c}
	\end{subfigure}

	\caption{\textbf{Visual Comparison of MCMarg and MCMarg-C Clustering Result.} In each row, there is a separate control group. On the left side are the visual results of MCMarg, while on the right side are the visual results of MCMarg-C. Each cluster is represented by points of different colors. The pie chart illustrates the proportion of different points in the overall distribution. We can observe that MCMarg-C exhibits a more uniform clustering pattern, while MCMarg tends to use a smaller number of Gaussians to describe the data distribution.}
	\label{fig:images}
\end{figure}

In the comparisons, we find in all datasets, MCMarg-C marginally outperforms others. 
In this experiment, we utilized the EM algorithm implemented in scikit-learn \cite{scikit-learn}.
This experiment also confirms the excellent distribution learning capabilities of the EM algorithm in low-dimensional spaces, which is widely recognized in the academic community. 
Given a situation that is beneficial to EM, MCMarg-C achieves even better results.
Considering that MCMarg-C can directly learn high-dimensional distributions and is differentiable, MCMarg-C may be the best distribution learning method in the field of clustering.

Table \ref{tab:integrated_comparison} presents a quantitative comparison between MCMarg-C and the original MCMarg \cite{dong2023bridging}.
We can see that the performance gap between MCMarg-C and MCMarg is even bigger than EM.
The advantages of MCMarg-C come from two factors.
First, MCMarg-C incorporates the GMM-Weight Standard Deviation Loss, which prevents clusters from dominating each other.
Second, MCMarg-C utilizes k-means as a prior to initialize the GMM means, which accelerates the convergence of distribution learning.
Thus, the improved MCMarg-C can achieve better performance compared to our original MCMarg.

To further validate the above statement, we generate images using Dalle-3 \cite{betker2023improving} and obtain their two-dimensional data points based on the grayscale values. 
Then, we perform distribution learning using both the MCMarg-C and MCMarg methods. 
The results are shown in Figure \ref{fig:images}. 
Each row in the figure represents a separate group. 
Different colors are used to represent each cluster with points. 
The pie chart illustrates the proportion of different points in the overall distribution. 
We can observe that MCMarg-C exhibits a more uniform clustering trend, while MCMarg tends to use a smaller number of Gaussians to describe the data distribution. 
For a clustering problem, MCMarg-C's performance is noticeably better.\\\\
\textbf{Further Discussion}

\begin{figure}[t]
	\includegraphics[width=\textwidth]{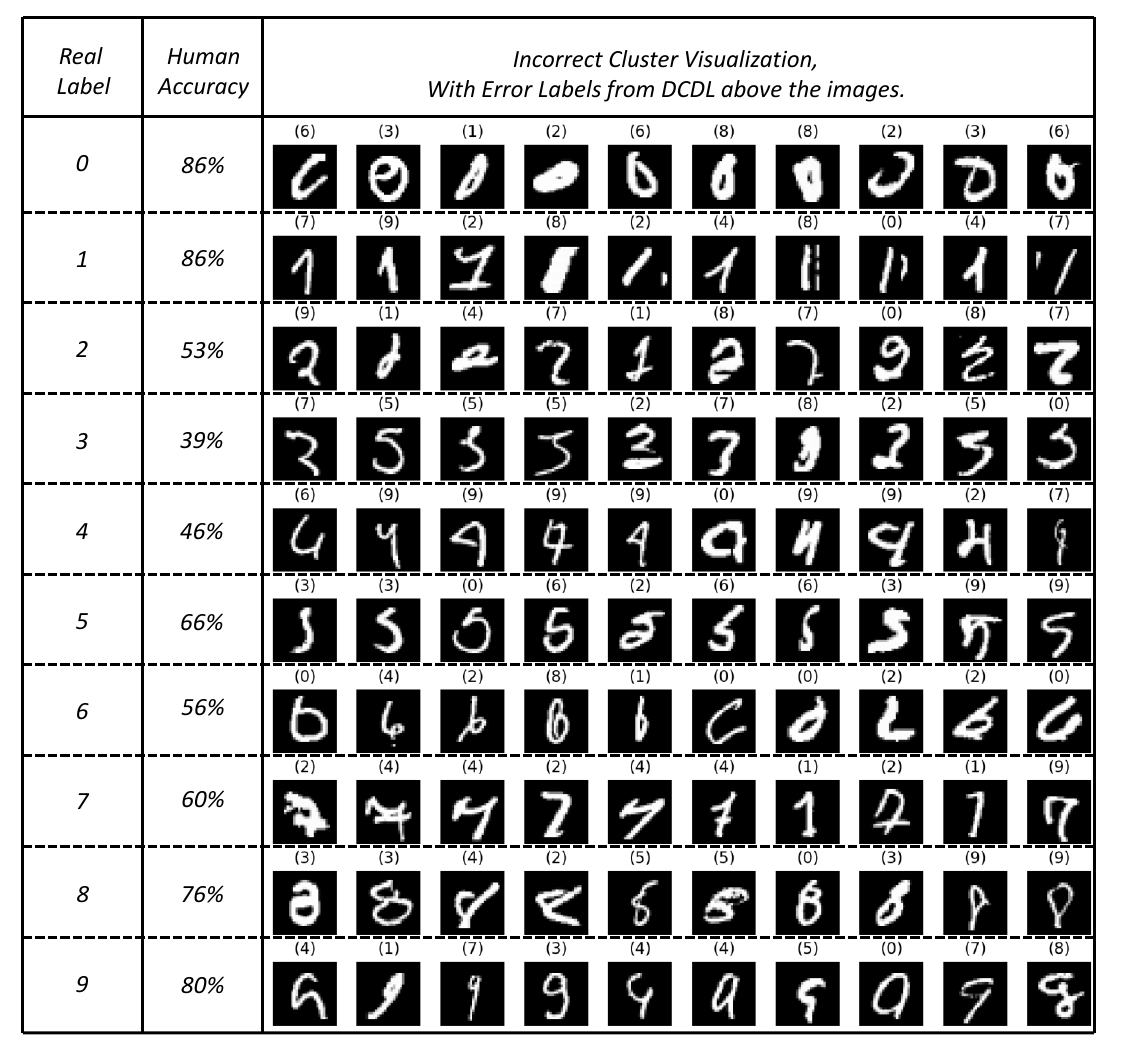}
	\caption{\label{failure_case}
		\textbf{DCDL Error Cluster Examples on the MNIST Dataset.} Real Label represents the true label of the images on the right. Incorrect Cluster Visualization shows the visual results of mis-clustered examples. The label results of DCDL are shown above each image. For Human Accuracy, we sought annotations from three individuals considering randomized image presentation. Accuracy reflects the agreement between human annotations and the ground truth labels in the dataset.}
\end{figure}

Figure \ref{failure_case} presents incorrect clustering examples by DCDL on the MNIST dataset. 
The Incorrect Cluster Visualization displays the misclassified examples. 
The numbers on each image represent DCDL's clustering labels. 
Additionally, we conduct a subjective study by three individuals with a background in computer vision to manually annotate these images. 
Before manual annotation, we shuffle the order of the images. 
Finally, we compute the accuracy of manual annotations compared to the ground truth labels. 
The results show that we do not achieve particularly high accuracy. The average accuracy among the three individuals was 65\%.

Through the observation of misclassified images, we find that these images are indeed prone to confusion. 
For instance, digits belonging to category `3' achieve only 39\% correct classification.
For the digit `3', misclassified samples tend to share a similar appearance with other digits, which is one of the reasons for the ineffectiveness of DCDL.
This finding shows the potential for DCDL in detecting mis-labeled data.

\section{Conclusion}
In this paper, we introduced a novel deep clustering algorithm called Deep Clustering via Distribution Learning (DCDL). 
This algorithm combines distribution learning with a deep clustering framework. 
With the theoretical analysis that supports distribution learning in clustering, we proposed clustering-optimized Monte Carlo Marginalization for clustering (MCMarg-C) to obtain clustering labels. 
Through the theoretical analysis and the improvements of MCMarg specifically designed for clustering, DCDL demonstrated superior performance compared to EM-based, VAE-based, and GAN-based distribution learning deep clustering methods.

\section*{Declarations}

\subsection*{Competing Interests}
This work is supported by NSERC Canada, and Alberta Innovates Discovery Supplement Grants.
This support has no influence on the design, execution, or interpretation of the present study.

\subsection*{Authors Contribution Statement}
Guanfang Dong, Zijie Tan, and Chenqiu Zhao contributed equally to the experimental design, code implementation, experiment observation, experiment recording, experiment analysis, and manuscript writing. Anup Basu supervised the entire process of the experiments and manuscript writing, providing revisions and experimental suggestions.

\subsection*{Ethical and Informed Consent for Data Used}
This user study was conducted following ethical guidelines. All participants provided informed consent before participating in the study.

\subsection*{Data Availability and Access}
The data used for this study will be made publicly available on GitHub upon the publication of this paper. A link to the repository will be provided at that time.

\bibliography{main}


\end{document}